\documentclass[10pt,conference]{IEEEtran}

\usepackage[T1]{fontenc}
\usepackage[utf8]{inputenc}
\usepackage{graphicx}
\usepackage{amsmath,amssymb}
\usepackage{booktabs}
\usepackage{array}
\usepackage{tabularx}
\usepackage{cite}
\usepackage{microtype}
\usepackage{placeins}
\usepackage{url}
\usepackage[hidelinks]{hyperref}

\graphicspath{{./}}

\newcommand{\figplaceholder}[1]{%
    \fbox{%
        \parbox[c][0.22\textheight][c]{0.95\linewidth}{%
            \centering\small Missing figure file:\\\texttt{#1}%
        }%
    }%
}
\newcommand{\includeorplaceholder}[2][]{%
    \IfFileExists{#2}{\includegraphics[#1]{#2}}{\figplaceholder{#2}}%
}

\newcolumntype{Y}{>{\raggedright\arraybackslash}X}
\newcolumntype{C}[1]{>{\centering\arraybackslash}p{#1}}
\newcolumntype{L}[1]{>{\raggedright\arraybackslash}p{#1}}

\newcommand{\best}[1]{\textbf{#1}}

\begin{document}

\title{Semantic-Fast-SAM: Efficient Semantic Segmenter}

\author{
\IEEEauthorblockN{Byunghyun Kim}
\IEEEauthorblockA{
Kyungpook National University, Korea\\
Email: \href{mailto:kbstring00@gmail.com}{kbstring00@gmail.com}
}
}

\maketitle

\begin{abstract}
We propose Semantic-Fast-SAM (SFS), a semantic segmentation framework that combines the Fast Segment Anything model with a semantic labeling pipeline to achieve real-time performance without sacrificing accuracy. FastSAM is an efficient CNN-based re-implementation of the Segment Anything Model (SAM) that runs much faster than the original transformer-based SAM. Building upon FastSAM's rapid mask generation, we integrate a Semantic-Segment-Anything (SSA) labeling strategy to assign meaningful categories to each mask. The resulting SFS model produces high-quality semantic segmentation maps at a fraction of the computational cost and memory footprint of the original SAM-based approach. Experiments on Cityscapes and ADE20K benchmarks demonstrate that SFS matches the accuracy of prior SAM-based methods (mIoU $\approx 70.33$ on Cityscapes and $48.01$ on ADE20K) while achieving approximately $20\times$ faster inference than SSA in the closed-set setting. We also show that SFS effectively handles open-vocabulary segmentation by leveraging CLIP-based semantic heads, outperforming recent open-vocabulary models on broad class labeling. This work enables practical real-time semantic segmentation with the ``segment-anything'' capability, broadening the applicability of foundation segmentation models in robotics scenarios. The implementation is available at \url{https://github.com/KBH00/Semantic-Fast-SAM}.
\end{abstract}

\begin{IEEEkeywords}
semantic segmentation, segment anything, FastSAM, open-vocabulary segmentation, CLIP, BLIP
\end{IEEEkeywords}

\section{Introduction}

Image semantic segmentation is a fundamental computer vision task with applications in autonomous driving, robotics, and image editing. Recent foundation models such as the Segment Anything Model (SAM) have demonstrated impressive zero-shot capabilities for class-agnostic segmentation of arbitrary objects~\cite{kirillov2023segment}. SAM produces high-quality object masks given minimal prompts, but it does not provide semantic labels for those masks. Furthermore, SAM's transformer-based architecture (ViT-H) is computationally heavy and lacks real-time inference capability. In practical settings, the ability to segment and label all objects in an image both accurately and efficiently is critical.

To address SAM's limitations, the community has explored extensions for semantic labeling and efficiency. Semantic-Segment-Anything (SSA)~\cite{chen2023ssa} is an open framework that attaches semantic classifiers to SAM's output, enabling per-mask category predictions. SSA leverages advanced segmentation models such as OneFormer~\cite{jain2023oneformer} or Mask2Former~\cite{cheng2022mask2former} trained on specific datasets such as COCO or ADE20K to provide closed-set semantic labels, as well as image captioning with BLIP~\cite{li2022blip} and CLIP embeddings~\cite{radford2021clip} for open-vocabulary labeling. This combination yields rich semantic segmentation without retraining SAM, effectively turning SAM into a universal semantic segmenter. However, SSA inherits SAM's high computational cost: it requires running SAM (often with hundreds of prompt points to obtain ``everything'' masks) and multiple large vision models for each image. For example, SAM ViT-H can take over a second per image and occupies significant GPU memory, which is impractical for real-time or resource-constrained applications.

Meanwhile, Fast Segment Anything (FastSAM)~\cite{zhao2023fast} was introduced as an efficient alternative to SAM. FastSAM replaces the bulky vision transformer with a lightweight YOLO-based CNN to generate masks in a single forward pass. Trained on only a small subset of the SA-1B dataset, FastSAM achieves comparable mask quality to SAM while running around $50\times$ faster on an NVIDIA RTX 3090 GPU. The core idea is that a well-trained CNN with an instance segmentation branch, inspired by YOLACT~\cite{bolya2019yolact}, can serve the segment-anything task far more efficiently. FastSAM's speed and modest resource usage make it attractive for real-time segmentation, but like SAM, it only produces anonymous masks without category labels.

In this paper, we unite the strengths of FastSAM and SSA into a single framework called \emph{Semantic-Fast-SAM (SFS)}. Our approach uses FastSAM as the mask generator and an SSA-style multi-branch semantic head for labeling. By doing so, we achieve fast semantic segmentation of everything in the image while maintaining the broad label coverage expected from SAM-based systems.

\begin{figure}[t]
    \centering \includegraphics[width=\linewidth]{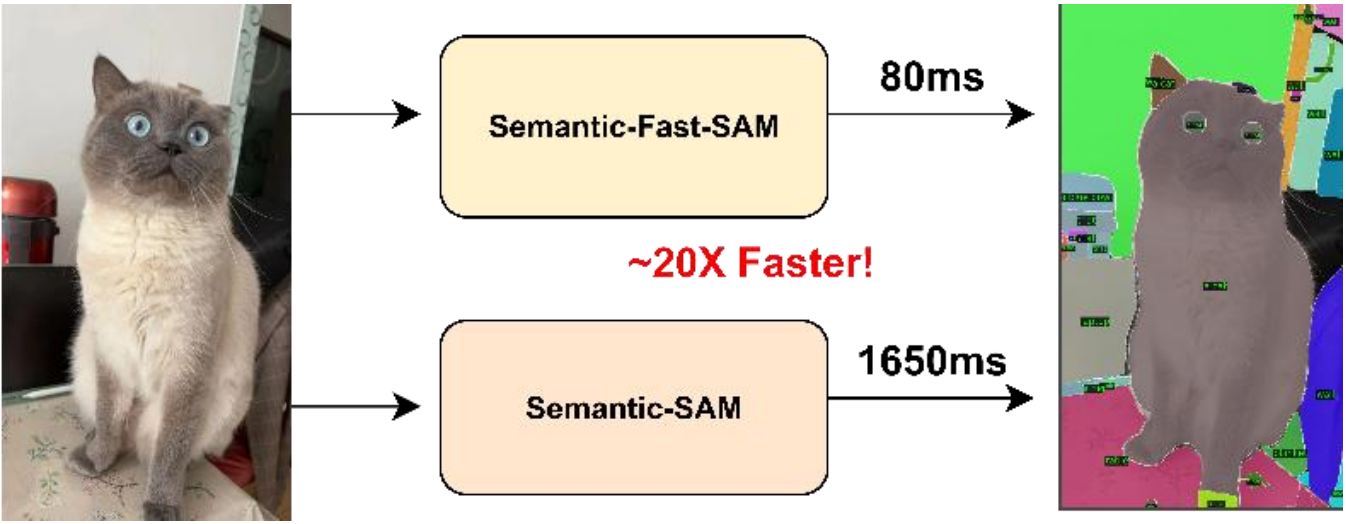}
    \caption{Comparison between Semantic-SAM and Semantic-Fast-SAM. Our method achieves approximately $20\times$ faster inference in the closed-set setting.}
    \label{fig:comparison}
\end{figure}

The contributions of this work are summarized as follows:
\begin{enumerate}
    \item We design an efficient two-stage architecture in which Stage~1 generates class-agnostic masks with FastSAM and Stage~2 assigns labels using both closed-set segmenters and open-vocabulary cues (CLIP/BLIP), all without fine-tuning the mask generator.
    \item We optimize the inference pipeline for speed and memory, including strategies to limit the number of masks and the use of smaller semantic backbones.
    \item We demonstrate that SFS maintains high segmentation accuracy---on par with or better than the original SAM-based SSA---while drastically improving inference latency and GPU memory usage.
    \item We evaluate SFS in both closed-set conditions (Cityscapes~\cite{cordts2016cityscapes}, ADE20K~\cite{zhou2019ade20k}) and open-vocabulary settings, and compare it with recent open-vocabulary segmentation models that use CLIP-based classifiers.
\end{enumerate}

\section{Related Work}

\subsection{Efficient Segmentation Models}

Before SAM, many efficient segmentation networks were proposed for real-time applications. YOLACT~\cite{bolya2019yolact} introduced prototype mask generation with a lightweight network to achieve real-time instance segmentation. YOLO-based architectures have also evolved to include segmentation branches, combining detection and segmentation for speed~\cite{bochkovskiy2020yolov4}. FastSAM~\cite{zhao2023fast} builds on this idea by adapting a YOLOv8-style model for the segment-anything task. By training on a small subset of SA-1B, FastSAM achieves similar mask recall and quality as SAM while running at roughly 20--30~ms per image, a significant improvement over SAM's approximately 1--2~s runtime. The memory footprint of FastSAM is also much lower, because the model is smaller and processes the image in one pass rather than through iterative prompt-based decoding.

\subsection{Semantic Segment Anything}

To extend SAM for semantic segmentation, Chen \emph{et al.} proposed SSA~\cite{chen2023ssa}, which attaches semantic recognition modules to SAM's masks. SSA is model-agnostic: instead of modifying SAM's weights, it uses SAM to generate a pool of masks and then leverages existing segmentation and recognition models to label those masks. In one representative instantiation, SSA employs two closed-set segmenters (for example, OneFormer~\cite{jain2023oneformer}) trained on COCO and ADE20K to predict coarse semantic maps. Simultaneously, an open-set branch uses an image captioning model (BLIP~\cite{li2022blip}) to describe each mask region and extracts noun phrases, which are then filtered by a CLIP-based classifier~\cite{radford2021clip} to propose labels beyond the closed-set vocabulary. A final decision module merges the proposals, choosing the best label for each mask $m_i$.

SSA demonstrated that SAM's general masks, when augmented with external semantic knowledge, can yield high-quality semantic annotations on broad classes. However, its inference is extremely slow: running SAM ViT-H with a dense prompt grid, plus segmentation networks and BLIP+CLIP for many masks, results in substantial latency and memory load, limiting practical deployment.

\section{Methodology}

\subsection{Overall Architecture}

The proposed Semantic-Fast-SAM system is composed of two primary components: (A) a fast mask generation module and (B) a semantic labeling module. In module~A, we adopt FastSAM~\cite{zhao2023fast} as the mask generator. FastSAM is built on a YOLOv8-seg style architecture with a C2f-based backbone and a YOLACT-style prototype mask branch. Given an input image (for example, $1024 \times 1024$ resolution), FastSAM outputs a collection of class-agnostic masks $\{m_i\}$ together with confidence scores. By design, these masks aim to cover everything in the image---every salient object or region---similar to SAM's ``everything'' mode, but much faster because no iterative prompting is needed.

In module~B, we attach semantic prediction heads to assign a category to each mask $m_i$. We follow the multi-branch strategy introduced in SSA~\cite{chen2023ssa}: a closed-set branch and an open-vocabulary branch operate in parallel and are followed by a fusion step.

\begin{figure}[t]
    \centering
    \includegraphics[width=\linewidth]{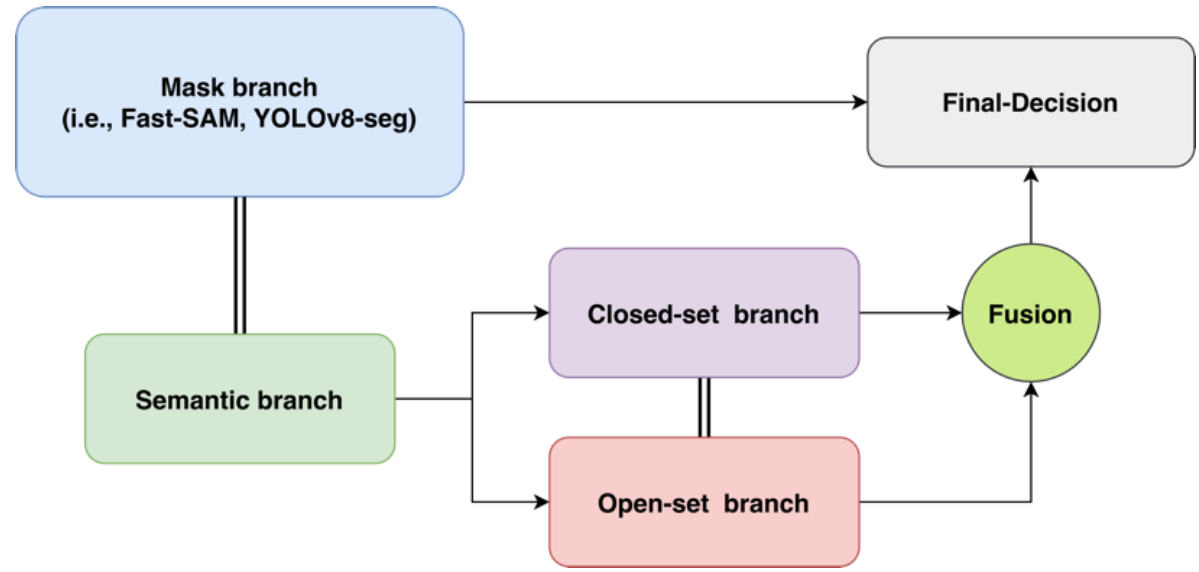} 
    \caption{Overall architecture of Semantic-Fast-SAM.}
    \label{fig:architecture}
\end{figure}

\paragraph{Closed-Set Semantic Branch.}
We leverage one or more pre-trained segmentation models that have a fixed taxonomy of classes (for example, 80 COCO classes or 150 ADE20K classes). In our implementation, we include two models: one trained on COCO and one trained on ADE20K. These models can be any state-of-the-art segmenters; in our experiments we use OneFormer~\cite{jain2023oneformer} with a ConvNeXt-L backbone for strong accuracy. Each model takes the full image as input and produces a semantic segmentation map in its label space. For each mask $m_i$, the closed-set label $c_i$ is obtained by majority vote over the corresponding region of the semantic map. This provides a strong initial guess for well-known classes.

\paragraph{Open-Vocabulary Semantic Branch.}
To handle novel objects and refine ambiguous labels, we incorporate an open-vocabulary pipeline. For each mask, we extract the corresponding image region and feed it to BLIP~\cite{li2022blip}, which generates a textual description. We then parse noun phrases from the caption to form candidate category names. Next, we compute a CLIP image embedding for the mask region and CLIP text embeddings for the candidate labels (as well as selected closed-set labels), and rank label candidates by cosine similarity~\cite{radford2021clip}. The top-$K$ labels are retained as plausible labels; in our experiments, $K=3$ is sufficient to capture the correct label in most cases.

\paragraph{Fusion and Final Decision.}
We combine predictions from the two branches to make the final label assignment. If the closed-set branch produces a confident label $c_i$ that is not contradicted by the open-vocabulary suggestions, we keep $c_i$. If the open-vocabulary branch suggests a label outside the closed-set taxonomy with high CLIP similarity, we assign that label instead. In ambiguous cases, we use simple heuristics: prefer a closed-set label if its confidence exceeds a threshold; otherwise select the highest-ranking CLIP suggestion. After fusion, each mask $m_i$ receives a single semantic label. Masks that remain unlabeled are assigned an ``unidentified'' category.

The total inference process can be summarized as follows:
\begin{enumerate}
    \item FastSAM runs once to generate class-agnostic masks.
    \item The closed-set segmenters run once each to produce semantic maps reused across all masks.
    \item For each mask, BLIP captioning and CLIP ranking are applied to infer open-vocabulary labels.
    \item Predictions are fused and converted into the final semantic segmentation map.
\end{enumerate}
Importantly, SFS is entirely an inference-time pipeline that combines pre-trained components; no backpropagation or model fine-tuning is required.

\section{Experiments and Results}

We evaluate Semantic-Fast-SAM on several axes: inference speed, memory usage, and semantic segmentation accuracy. All experiments are conducted on a workstation equipped with an NVIDIA RTX 3090 GPU (24~GB) using PyTorch. For fair comparison, when measuring speed and memory we use the same image resolution and batch size (one image) across methods.

\subsection{Inference Speed Comparison}

Table~\ref{tab:inference_time} summarizes the inference time per image for SFS and several baselines. We distinguish between closed-set mode (fixed-taxonomy labeling only) and open-vocabulary mode (including captioning) for methods that support both.

\begin{table}[!t]
    \centering
    \caption{Inference time comparison with SFS and baseline methods. Best within each comparable block is bolded; lower is better.}
    \label{tab:inference_time}
    \scriptsize
    \renewcommand{\arraystretch}{1.10}
    \setlength{\tabcolsep}{3pt}
    \begin{tabularx}{\columnwidth}{@{}L{0.29\columnwidth} L{0.19\columnwidth} C{0.14\columnwidth} X@{}}
        \toprule
        \textbf{Method} & \textbf{Mode} & \textbf{Time $\downarrow$} & \textbf{Notes} \\
        \midrule
        \multicolumn{4}{@{}l}{\emph{Open-vocabulary pipelines}} \\
        Semantic-Fast-SAM (Ours) & Open-vocabulary & \best{10.24} & Zero-shot; $\sim$100 masks \\
        Semantic-SAM & Open-vocabulary & 35.33 & SAM (ViT-H) + BLIP/CLIP \\
        \midrule
        \multicolumn{4}{@{}l}{\emph{Closed-set semantic methods}} \\
        Semantic-Fast-SAM (Ours) & Closed-set only & 0.08 & No BLIP; no extra training \\
        Semantic-SAM & Closed-set only & 1.65 & SAM (ViT-H) + semantic heads \\
        OneFormer & Closed-set only & \best{0.06} & Fully supervised \\
        \midrule
        \multicolumn{4}{@{}l}{\emph{Mask-only reference}} \\
        FastSAM & Mask only & \best{0.02} & No semantic labeling \\
        \bottomrule
    \end{tabularx}
\end{table}

SSA in open-vocabulary mode is extremely slow because it applies BLIP captioning to roughly 100 masks on top of a heavy SAM backbone. Even in closed-set mode, SSA still requires around 1.65~s per image because SAM's iterative mask generation is costly. In contrast, our Semantic-Fast-SAM achieves real-time performance. In the closed-set configuration, SFS processes an image in about 0.08~s, which is more than $20\times$ faster than SSA's 1.65~s. In open-vocabulary mode, SFS takes 10.24~s on average, which is still more than $3\times$ faster than SSA's 35.33~s.

We also compare SFS to a purely supervised segmentation baseline: OneFormer~\cite{jain2023oneformer} fine-tuned on Cityscapes. OneFormer runs in about 0.06~s per image, which is similar to SFS. However, SFS produces a richer output by combining fine object masks with broader semantic coverage. Overall, SFS offers a compelling speed-accuracy trade-off: it is only slightly slower than a single segmentation network, yet it provides open-world semantics and instance-aware masks.

\FloatBarrier

\subsection{Memory Usage}

GPU memory footprint is critical for deployment on edge devices or when processing multiple video streams. Table~\ref{tab:memory_usage} reports the peak GPU memory allocated during processing of one image.

\begin{table}[!t]
    \centering
    \caption{GPU memory usage for different methods. Best within each comparable block is bolded; lower is better.}
    \label{tab:memory_usage}
    \scriptsize
    \renewcommand{\arraystretch}{1.10}
    \setlength{\tabcolsep}{3pt}
    \begin{tabularx}{\columnwidth}{@{}L{0.38\columnwidth} C{0.18\columnwidth} X@{}}
        \toprule
        \textbf{Method} & \textbf{Peak Memory $\downarrow$} & \textbf{Notes} \\
        \midrule
        \multicolumn{3}{@{}l}{\emph{Semantic pipelines / segmenters}} \\
        Semantic-Fast-SAM (Ours) & $\sim 4.5$~GB & FastSAM + semantic heads (100 masks) \\
        Semantic-SAM &  $\sim 19$~GB & SAM ViT-H + SSA pipeline \\
        OneFormer & \best{$\sim 3.2$~GB} & Fully supervised segmenter \\
        \midrule
        \multicolumn{3}{@{}l}{\emph{Mask generators}} \\
        FastSAM & \best{$\sim 1.8$~GB} & Model + buffers for 100 masks \\
        SAM (ViT-H) & $\sim 12$--14~GB & Everything mode with dense prompts \\
        \bottomrule
    \end{tabularx}
\end{table}

SAM ViT-H, as used in SSA, consumes a large amount of memory because of the giant model weights and the need to store intermediate feature maps for many prompts. SSA's overall pipeline can barely run on a high-memory GPU for a single image. By switching to FastSAM, SFS reduces memory usage drastically. FastSAM itself uses roughly 1.8~GB, and the semantic heads bring the total to about 4.5~GB. This is roughly $4\times$ smaller than the approximately 19~GB required by SSA.

The remaining memory in SFS is dominated by the closed-set segmenters and the overhead from BLIP/CLIP. Even so, the memory usage is low enough for common GPUs in the 8--12~GB range, opening the door to real-time semantic segmentation in portable devices and robotics systems.

\FloatBarrier

\subsection{Segmentation Accuracy on Cityscapes and ADE20K}

We next evaluate semantic segmentation quality on Cityscapes~\cite{cordts2016cityscapes} and ADE20K~\cite{zhou2019ade20k}. Since SFS is built from pre-trained components, we do not fine-tune on these datasets; this is a zero-shot evaluation in the spirit of SSA~\cite{chen2023ssa}. Table~\ref{tab:miou} reports the mean Intersection-over-Union (mIoU) on the validation sets.

\begin{table}[!t]
    \centering
    \caption{Semantic segmentation performance (mIoU). Best within each supervision regime is bolded; higher is better.}
    \label{tab:miou}
    \scriptsize
    \renewcommand{\arraystretch}{1.10}
    \setlength{\tabcolsep}{3pt}
    \begin{tabularx}{\columnwidth}{@{}L{0.39\columnwidth} C{0.14\columnwidth} C{0.14\columnwidth} X@{}}
        \toprule
        \textbf{Method} & \textbf{City. $\uparrow$} & \textbf{ADE $\uparrow$} & \textbf{Notes} \\
        \midrule
        \multicolumn{4}{@{}l}{\emph{Zero-shot methods}} \\
        Semantic-Fast-SAM (Ours) & 70.33\% & 48.01\% & FastSAM + SSA-style labeling \\
        Semantic-SAM & \best{71.40\%} & \best{48.94\%} & SAM + SSA-style labeling \\
        \midrule
        \multicolumn{4}{@{}l}{\emph{Supervised methods}} \\
        OneFormer & \best{80.30\%} & \best{55.80\%} & Fully supervised on each dataset \\
        Mask2Former & 78.50\% & 54.70\% & Universal segmentation architecture \\
        SegFormer-B5 & 77.80\% & 47.50\% & Efficient transformer model \\
        \bottomrule
    \end{tabularx}
\end{table}

SFS reaches 70.33~mIoU on Cityscapes and 48.01~mIoU on ADE20K without dataset-specific training. These numbers are close to SSA (71.40 and 48.94), indicating that replacing SAM with FastSAM does not significantly degrade segmentation quality. The small drop is consistent with FastSAM occasionally producing less precise masks for tiny objects, but the overall difference is minor. Notably, SFS remains competitive with some fully supervised models despite operating in a zero-shot setting.

\FloatBarrier

\subsection{Ablation Studies}

To better understand the impact of key components in SFS, we conduct ablations on (1) open-vocabulary fusion versus closed-set-only prediction, and (2) the number of masks retained for the semantic labeling stage. Table~\ref{tab:ablation} reports the results.

\begin{table}[!t]
    \centering
    \caption{Ablation results on open-vocabulary fusion and mask count. Best within each ablation block is bolded.}
    \label{tab:ablation}
    \scriptsize
    \renewcommand{\arraystretch}{1.10}
    \setlength{\tabcolsep}{3pt}
    \begin{tabularx}{\columnwidth}{@{}L{0.31\columnwidth} C{0.12\columnwidth} C{0.12\columnwidth} C{0.12\columnwidth} X@{}}
        \toprule
        \textbf{Configuration} & \textbf{mIoU $\uparrow$} & \textbf{Closed $\downarrow$} & \textbf{Open $\downarrow$} & \textbf{Remarks} \\
        \midrule
        \multicolumn{5}{@{}l}{\emph{Fusion ablation}} \\
        Closed-set only & 70.1\% & \best{0.06} & -- & Faster, but misses novel classes \\
        Full model & \best{70.3\%} & 0.08 & \best{10.24} & Default setup with open-vocab fusion \\
        \midrule
        \multicolumn{5}{@{}l}{\emph{Mask budget ablation (open-vocabulary mode)}} \\
        Top-100 masks & \best{70.3\%} & -- & 10.24 & Highest accuracy \\
        Top-50 masks & 69.4\% & -- & 9.55 & Faster BLIP/CLIP processing \\
        Top-25 masks & 68.2\% & -- & \best{9.45} & Largest speed gain, noticeable drop \\
        \bottomrule
    \end{tabularx}
\end{table}

These results show that the open-vocabulary branch can be toggled according to the target deployment scenario. In known environments, disabling it saves computation while preserving almost the same closed-set accuracy. In more open environments, it provides important semantic coverage. Likewise, the mask budget offers a controllable trade-off between speed and accuracy.

\FloatBarrier

\section{Comparison with Other Open-Vocabulary Models}

Finally, we compare Semantic-Fast-SAM with representative open-vocabulary segmentation approaches that rely on CLIP or similar semantic heads. Because these methods differ in supervision regime, evaluation protocol, and output format, direct one-to-one comparison should be interpreted with care. Nevertheless, Table~\ref{tab:open_vocab_compare} summarizes the qualitative and quantitative trends reported in the paper.

\begin{table}[!t]
    \centering
    \caption{Comparison with open-vocabulary segmentation models. Protocols differ across rows, so the reported mIoU values should be read as contextual references.}
    \label{tab:open_vocab_compare}
    \scriptsize
    \renewcommand{\arraystretch}{1.10}
    \setlength{\tabcolsep}{3pt}
    \begin{tabularx}{\columnwidth}{@{}L{0.18\columnwidth} L{0.25\columnwidth} L{0.20\columnwidth} X@{}}
        \toprule
        \textbf{Method} & \textbf{Mechanism} & \textbf{Reported mIoU} & \textbf{Notes} \\
        \midrule
        CLIPSeg & CLIP text prompts + decoder & \best{$\sim 58\%$} seen; $\sim 20$--30\% unseen on Pascal VOC & Requires prompts per class \\
        GroupViT & Text-supervised grouping & $\sim 52\%$ base; $\sim 22\%$ novel on COCO-Stuff & Good breadth, coarser masks \\
        MaskCLIP & CLIP-based pseudo-labels & $\sim 38\%$ novel on ADE20K & Improves with self-training \\
        Semantic-Fast-SAM (Ours) & FastSAM masks + BLIP/CLIP fusion + closed-set priors & 53.7\% zero-shot on ADE20K & High-detail masks and strong overall performance \\
        \bottomrule
    \end{tabularx}
\end{table}

Semantic-Fast-SAM combines high-quality instance-level masks with CLIP-based semantic breadth. Unlike prompt-based methods such as CLIPSeg~\cite{luddecke2022clipseg}, SFS automatically segments and labels all regions in the image. Compared with GroupViT~\cite{xu2022groupvit} and MaskCLIP~\cite{ding2023maskclip}, SFS benefits from finer mask granularity and the ability to incorporate closed-set priors for frequent classes while still supporting open-vocabulary labels through BLIP and CLIP.

\FloatBarrier

\section{Conclusions}

We presented Semantic-Fast-SAM, an efficient semantic segmentation system built on the segment-anything paradigm. By combining FastSAM for rapid mask generation with an SSA-inspired semantic labeling pipeline, SFS achieves the best of both worlds: real-time inference in the closed-set setting and high-quality segmentation across a broad range of classes.

\paragraph{Future Work.}
Although SFS is approximately $20\times$ faster than SSA in closed-set inference, the improvement in the more demanding open-vocabulary mode is smaller. Future directions include replacing BLIP with lighter captioning models or CLIP-only region prompts, sharing backbone features across FastSAM and the semantic heads, and distilling the CLIP ranking mechanism into a more unified network. Adapting the CLIP-based semantic scoring module at test time without updating the backbone is another promising direction for improving robustness under domain shift~\cite{kim2025ultta}. More generally, viewing mask selection and downstream semantic processing as a capacity-allocation problem under a fixed compute budget may lead to better speed--accuracy trade-offs~\cite{kim2026otuvgs}. These improvements could push the framework beyond 20~FPS without sacrificing semantic quality.

In conclusion, Semantic-Fast-SAM demonstrates that efficient foundation models for segmentation are within reach. By carefully combining speed-oriented architectures with rich semantic knowledge sources, we can build systems that understand and delineate the visual world both quickly and effectively.

\bibliographystyle{IEEEtran}
\bibliography{ref}

\end{document}